\documentclass[runningheads]{llncs}

\usepackage[pdftex]{graphicx}
\usepackage[longtable]{multirow}
\usepackage{longtable}
\usepackage{epstopdf}
\usepackage{xcolor}
\usepackage{lipsum}
\usepackage{authblk}
\usepackage{amsmath,amssymb,mathtools,bm,etoolbox}
\usepackage{algorithmic}
\usepackage[ruled,vlined]{algorithm2e}
\usepackage[super]{nth}
\usepackage[section]{placeins}
\usepackage[hyphens]{url}
\usepackage{booktabs}
\usepackage{siunitx}
\usepackage{multirow}
\usepackage{subfigure}
\usepackage{rotating}
\usepackage{mwe}
\usepackage{hyperref}
\newcolumntype{L}{>{\centering\arraybackslash}m{2cm}}
\usepackage[normalem]{ulem}
\usepackage[misc]{ifsym}

\begin{document}
\title{Edge-Based Self-Supervision for Semi-Supervised Few-Shot Microscopy Image Cell Segmentation}

\author{Youssef Dawoud \inst{1} (\Letter) \and
Katharina Ernst\inst{2} \and
Gustavo Carneiro\inst{3} \\ \and
Vasileios Belagiannis\inst{4}}

% index{Dawoud, Youssef} 
% index{Ernst, Katharina} 
% index{Carneiro, Gustavo} 
% index{Belagiannis, Vasileios} 

\authorrunning{Y. Dawoud et al.}
\titlerunning{Edge-Based Supervision for Semi-Supervised Few-Shot Cell Segmentation}

\institute{Universit\"at Ulm, Ulm, Germany  \and Ulm University Medical Center, Ulm, Germany \and The University of Adelaide, Adelaide, Australia \and Otto von Guericke University, Magdeburg, Germany \\ \email{firstname.lastname@\{uni-ulm.de, adelaide.edu.au, ovgu.de\}}}
\maketitle              
\begin{abstract}
Deep neural networks currently deliver promising results for microscopy image cell segmentation, but they require large-scale labelled databases, which is a costly and time-consuming process. In this work, we relax the labelling requirement by combining self-supervised with semi-supervised learning. We propose the prediction of edge-based maps for self-supervising the training of the unlabelled images, which is combined with the supervised training of a small number of labelled images for learning the segmentation task. In our experiments, we evaluate on a few-shot microscopy image cell segmentation benchmark and show that only a small number of annotated images, e.g. 10\% of the original training set, is enough for our approach to reach similar performance as with the fully annotated databases on 1- to 10-shots. Our code and trained models is made publicly available \url{https://github.com/Yussef93/EdgeSSFewShotMicroscopy}.
\end{abstract}
\section{Introduction}
\label{sec:intro}

%\cite{ernst2018pharmacological,markey2019assays}
The analysis of microscopy images is usually focused on cell detection, counting, and segmentation. For instance, the analysis of changes in cell number and morphology induced by bacterial protein toxins contributes to assessing their activity~\cite{ernst2018pharmacological}, where these toxins are the causative agents of severe diseases like diphtheria, anthrax or whooping cough. Moreover, the cell number and morphology is used to assess residual activity of inactivated toxins in vaccines or to investigate inhibitors of toxins in order to develop novel therapeutic strategies \cite{ernst2018pharmacological,markey2019assays}. Data-driven approaches, such as deep neural networks~\cite{ciresan2012deep,long2015fully}, represent a key contribution towards reliably automating the tasks of cell detection, counting and segmentation \cite{xing2016robust}. However, one major drawback associated with deep neural networks is their dependence on large-scale labelled data sets for learning a fully-supervised model. This requires exhaustive manual labelling for every new microscopy data set on a pixel-level basis. 

\begin{figure}[ht!]
\centering

    \rotatebox[origin=l]{90}{TNBC}
	\includegraphics[width=3.5cm]{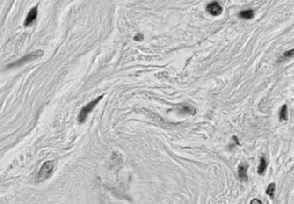}
	\includegraphics[width=3.5cm]{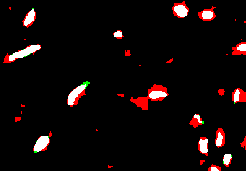}
	\includegraphics[width=3.5cm]{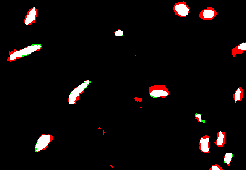}
	
	\rotatebox[origin=l]{90}{ssTEM }
	\includegraphics[width=3.5cm]{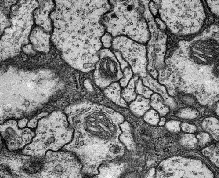}
	\includegraphics[width=3.5cm]{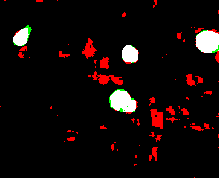}
	\includegraphics[width=3.5cm]{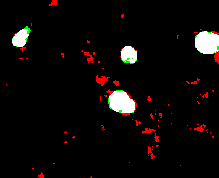}
	
	\rotatebox[origin=l]{90}{EM }
	\subfigure[c][Input]{\includegraphics[width=3.5cm]{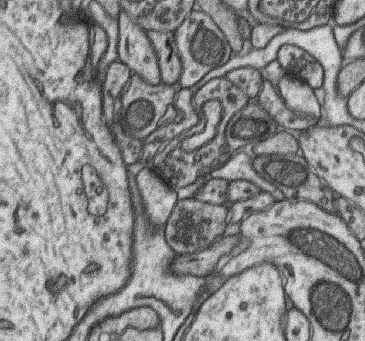}}
	\subfigure[c][$10\% \: \mathcal{S}^{L}$]{\includegraphics[width=3.5cm]{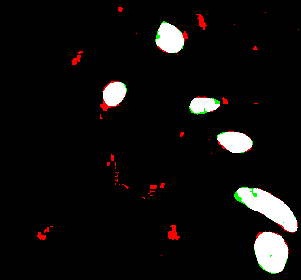}}
	\subfigure[c][$10\% \: \mathcal{S}^{L}+30\% \: \mathcal{S}^U$]{\includegraphics[width=3.5cm]{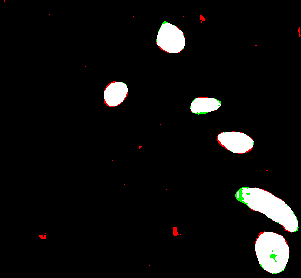}}

	\caption{Visual Result. We visually compare the joint training of our proposed edge-detection proxy task using labelled images plus 30\% of unlabelled images (i.e. $10\% \: \mathcal{S}^{L}+30\% \: \mathcal{S}^U$) to a supervised model trained on the labelled images only (i.e. $10\% \: \mathcal{S}^{L}$). The red color corresponds to false positive, the green color to false negative, the black color to true negative, and the white color to true positive. Best viewed in color.}
	\label{fig:VisualComp1}
\end{figure}

\begin{figure}[t]
	\centering
	\includegraphics[width=0.8\textwidth]{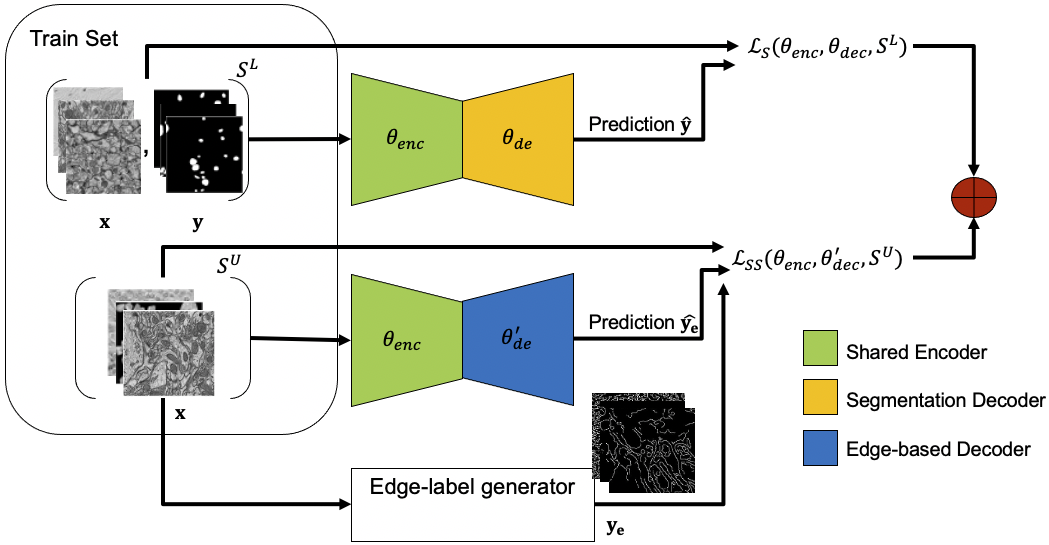}
   
	\caption{An overview of our learning algorithm. We perform joint training with both labelled set $\mathcal{S}^{L}$ and unlabelled images $\mathcal{S}^{U}$, where we utilize a Canny edge filter \cite{ding2001canny} to generate edge-based supervision using the unlabeled images in $\mathcal{S}^{U}$. }
	 \label{figtraining}
\end{figure}

Self-supervised and semi-supervised learning approaches target addressing the problem of limited available labels in different ways. In self-supervised approaches, the supervision stems from the data itself by learning proxy tasks such as rotation prediction \cite{gidaris2018unsupervised} or using contrastive learning \cite{chen2020simple}. Semi-supervised approaches work with a small portion of labelled data, while there are label-free objective functions for the unlabelled part, e.g.~entropy minimization \cite{grandvalet2005semi} and consistency enforcing \cite{scribble2label,cheplygina2019not}, similar to self-supervision. In medical image analysis, JigSaw \cite{noroozi2016unsupervised} and rotation proxy tasks are typical for self-supervision in 3D computed tomography (CT) scans \cite{taleb20203d}. Furthermore, brain segmentation from MRI images can be accomplished with a limited number of labelled data~\cite{cui2019semi}.

%Paragraph3 Our contribution
In this paper, we relax the extensive labelling requirement for cell segmentation. Motivated by self-supervision, we leverage unlabelled images of different microscopy databases to extract edge-based image features, which we propose as a new proxy task for training the deep neural network. Next to training for predicting image edges, we also assume a small number of labelled images for learning the segmentation task, as in semi-supervised learning. We evaluate our approach on the few-shot microscopy image cell segmentation benchmark by Dawoud \textit{et al.}~\cite{Dawoud2020}. To the best of our knowledge, this is the first work on edge-based self-supervision for semi-supervised microscopy cell image few-shot segmentation.

\section{Method}

\subsection{Problem Definition}

We consider the collection of data sets $\mathcal{S} = \{ \mathcal{S}_{1}, \mathcal{S}_{2},\dots, \mathcal{S}_{|\mathcal{S}|}\}$. Each data set $\mathcal{S}_{m}=\{ \mathcal{S}^{L}_{m},\mathcal{S}^{U}_{m}\}$ consists of the labelled subset of microscopy images $\mathcal{S}^{L}_{m}$ and the unlabelled image subset $\mathcal{S}^{U}_{m}$ respectively. Moreover, each dataset consists of different microscopy image type and cell segmentation task. The labelled subset is defined as $\mathcal{S}^{L}_{m}=\{(\mathbf{x},\mathbf{y})_{k}\}_{k=1}^{|\mathcal{S}^{L}_{m}|}$, where $(\mathbf{x},\mathbf{y})_{k}$ is a pair of the microscopy image $\mathbf{x}$ and pixel-level binary annotation $\mathbf{y}$. Moreover, the unlabelled subset $\mathcal{S}^{U}_{m}=\{(\mathbf{x})_k\}_{k=1}^{|\mathcal{S}^{U}_{m}|}$ contains a significant larger number of images such that $|\mathcal{S}^{U}_{m}| >> |\mathcal{S}^{L}_{m}|$.

Our objective is to learn a generic segmentation model $\mathbf{\hat{y}}=f(\mathbf{x};\theta)$, represented by a deep neural network with parameters $\theta$, by leveraging all information of $\mathcal{S}$. Afterwards, we aim to segment cells on the target data set $\mathcal{T}$ with the generic model. However, the microscopy images in $\mathcal{T}$ differ from $\mathcal{S}$ and thus we should adapt the segmentation model. This is a common situation in actual microscopy problems. For this reason, we assume access to a small annotated image set in the target set $\tilde{\mathcal{T}} \subset \mathcal{T}$, e.g.~5 images, which we can use for fine-tuning our model. Under this definition, we target a semi-supervised few-shot microscopy image cell segmentation problem with a self-supervised learning algorithm. Next, we present our approach to a generic segmentation model with edge-based self-supervision and a small number of annotated images.

\subsection{Learning Algorithm}

Edge extraction is a well-suited task for microscopy images given the shape and form of the cells. We propose an edge-based objective function for the unlabelled subset $\mathcal{S}^{U}_{m}$. To that end, we employ the Canny-edge detector~\cite{ding2001canny} to create an edge map for each unlabelled microscopy image. The updated subset $\mathcal{S}^{U}_{m} =\{(\mathbf{x},\mathbf{y}_{e})_k\}_{k=1}^{|\mathcal{S}^{U}_{m}|}$ now contains a binary edge map $\mathbf{y}_{e}$ for each microscopy image $\mathbf{x}$, which acts as a proxy task to learn the model parameters. In practice, we observed that it is not meaningful to completely rely on the same model for cell segmentation and edge prediction. For this reason, we decompose the segmentation model $f(\cdot)$ into the encoder and decoder parts, which are parametrized by $\theta_{en}$ and $\theta_{de}$, where $\theta =\{\theta_{en}, \theta_{de} \}$. Then, we design a second encoder-decoder model for edge prediction that shares the same encoder with the cell segmentation model. To train the edge prediction deep neural network, i.e.~$\theta^{\prime} =\{\theta_{en}, \theta^{\prime}_{de} \}$, we define the the self-supervised objective as following:

\begin{equation}\label{bce_canny}
    \begin{aligned}
     \mathcal{L}_{SS} (\theta_{en}, \theta_{de}^{\prime}, \mathcal{S}^{U}) &=
         - \frac{1}{|\mathcal{S}^{U}|} \sum_{ \mathcal{S}^{U}_{m}\in \mathcal{S}^{U}} \sum_{(\mathbf{x}, \mathbf{y}_{e})\in \mathcal{S}^{U}_{m}} \sum_{\omega \in \Omega }^{ }\left[\mathbf{w}(\omega) \mathbf{y}_{e}(\omega)\log( \mathbf{\hat{y}}_{e}(\omega) ) +   \right. \\
         &\left. (1-\mathbf{y}_{e}(\omega)) \log(1-\mathbf{\hat{y}}_{e} (\omega)) 
         \right] ,
\end{aligned}
\end{equation}

where $\mathcal{S}^{U} = \{ \mathcal{S}^{U}_{1}, \mathcal{S}^{U}_{2},\dots, \mathcal{S^{U}}_{|\mathcal{S^{U}}|}\}$ corresponds to all unlabelled subsets, $\omega \in \Omega$ denotes the spatial pixel position in the image lattice $\Omega$ and $\mathbf{\hat{y}}_{e}(\omega)$ represents the output of edge prediction model at the pixel location $(\omega)$. The proposed objective is a pixel-wise binary-cross entropy scaled by foreground weighting factor $\mathbf{w}(\omega)$, which equals to the ratio of background to foreground classes in $\mathbf{y}_{e}$. Note that it is important to use the weighting factor because of the imbalanced foreground to background ratio in the data sets.

Apart from the unlabelled data, we also use the labelled microscopy images of each data set, which we group as $\mathcal{S}^{L} = \{ \mathcal{S}^{L}_{1}, \mathcal{S}^{L}_{2},\dots, \mathcal{S}^{L}_{|\mathcal{S}^{L}|}\}$. Similarly to Eq.~\ref{bce_canny}, we define the objective as the weighted binary cross entropy, given by:  
\begin{equation}\label{bce_seg}
    \begin{aligned}
     \mathcal{L}_{S} (\theta_{en}, \theta_{de}, \mathcal{S}^{L}) &=
         - \frac{1}{|\mathcal{S}^{L}|} \sum_{ \mathcal{S}^{L}_{m}\in \mathcal{S}^{L}} \sum_{(\mathbf{x}, \mathbf{y})\in \mathcal{S}^{L}_{m}} \sum_{\omega \in \Omega }^{ }\left[\mathbf{w}(\omega) \mathbf{y}(\omega)\log( \mathbf{\hat{y}}(\omega) ) +   \right. \\
         &\left. (1-\mathbf{y}(\omega)) \log(1-\mathbf{\hat{y}} (\omega)) 
         \right] ,
\end{aligned}
\end{equation}
where $\mathbf{\hat{y}}(\omega)$ is the output of the segmentation model at the pixel location $\omega$, while the weighing scheme is identical to Eq.~\ref{bce_canny}. Finally, the parameters of segmentation and edge prediction models are jointly learned based on Eq.~\ref{bce_seg} and Eq.~\ref{bce_canny}. Our complete objective is described by: 
\begin{equation}\label{JointOpt-eq}
\arg \min_{\theta_{en}, \theta_{de}, \theta_{en}^{\prime}}  [\mathcal{L}_{SS} (\theta_{en}, \theta_{de}^{\prime}, \mathcal{S}^{U}) + \mathcal{L}_{S} (\theta_{en}, \theta_{de}, \mathcal{S}^{L})],
\end{equation}
where the minimization is accomplished with backpropagation and mini-batch stochastic gradient descent for the parameter update. Our learning algorithm is illustrated in Fig.~\ref{figtraining}.

We only keep the segmentation model after completing the joint learning. The last step of our approach is to fine-tune the segmentation model with the few annotated images i.e. $K$-shots from the target data set $\tilde{\mathcal{T}}$. This optimization is described as: 
\begin{equation}
    \theta^* = \arg \min_{\theta_{enc},\theta_{de}} [\mathcal{L}_{S}(\theta_{enc},\theta_{de},\tilde{\mathcal{T}})].
\end{equation}
Finally, the fine-tuned model with the updated parameters $\theta^*$ is evaluated on the target test set $\hat{\mathcal{T}} = \mathcal{T} \setminus \tilde{\mathcal{T}}$.

\paragraph{Implementation} We rely on the encoder-decoder model of the fully convolutional regression network (FCRN) from \cite{xie2018microscopy}. Additionally, we implement the loss functions for each mini-batch. We train with two Adam optimizers to update each decoder separately and apply the same model modifications and hyper-parameters, as in \cite{Dawoud2020}. 

\section{Experiments}
We perform our evaluation based on the protocol from~\cite{Dawoud2020}. In particular, we use the B5 and B39 data sets from the Broad Bioimage Benchmark Collection (BBBC)~\cite{lehmussola2007computational}. The former contains 1200 fluorescent synthetic stain cells images, while the latter contains 200 fluorescent synthetic stain cells. Second, we have ssTEM~\cite{gerhard2013segmented} and EM~\cite{lucchi2013learning} data sets which contains 165 and 20 electron microscopy images respectively of mitochondria cells. At last, the TNBC data set consists of 50 histology images of breast biopsy~\cite{naylor2018segmentation}. 

We split each data set to 10\% labelled and use the rest as unlabelled, similar to semi-supervision protocol in~\cite{sohn2020simple}. We study the effect of jointly learning the edge prediction and cell segmentation tasks on the overall segmentation performance after fine-tuning on the $K$-shots and testing on the test set of the target data set. To this end, we first train a fully-supervised model on $10\%$ of $\mathcal{S}^L$, then we incrementally add 30\%, 60\%, and 100\% from $\mathcal{S}^{U}$. We train for 50 epochs with batch size of 64 and Adam optimizer with 0.001 learning rate. We also compare our approach against entropy \cite{grandvalet2005semi} and consistency regularization \cite{scribble2label}. Furthermore, we compare our approach of edge-detection as self-supervised task against the self-supervised contrastive learning (SimCLR) \cite{chen2020simple} and rotation prediction \cite{gidaris2018unsupervised}. We adapt SimCLR and rotation predicition approaches to the microscopy image domain by pre-training the encoder on all the images in $\mathcal{S}^{U}$. For SimCLR, we pre-train the encoder using the contrastive loss for 200 epochs with a batch size of 1024, an Adam optimizer with 0.003 learning rate and cosine annealing scheduler. As for rotation prediction  we pre-train the encoder to classify the rotation degree i.e. images which are rotated with either 0°, 90°, 180°, or 270°. Moreover, we train for 50 epochs with a batch size of 64 and an SGD optimizer with 0.1 learning rate. Afterwards, a decoder is trained on top of the pre-trained encoders from SimCLR and rotation on the cell segmentation task using $\mathcal{S}^L$. Finally, all the pre-trained models are fine-tuned on $\tilde{\mathcal{T}}$ and tested on $\hat{\mathcal{T}}$. The mean intersection over union (IoU) is computed over 10 random selections of 1-,3-,5-,7-,10-shots.

\subsection{Results and Discussion}

\begin{figure}[ht!]
    \centering
    \subfigure[1-shot.]{\includegraphics[width=0.45\textwidth]{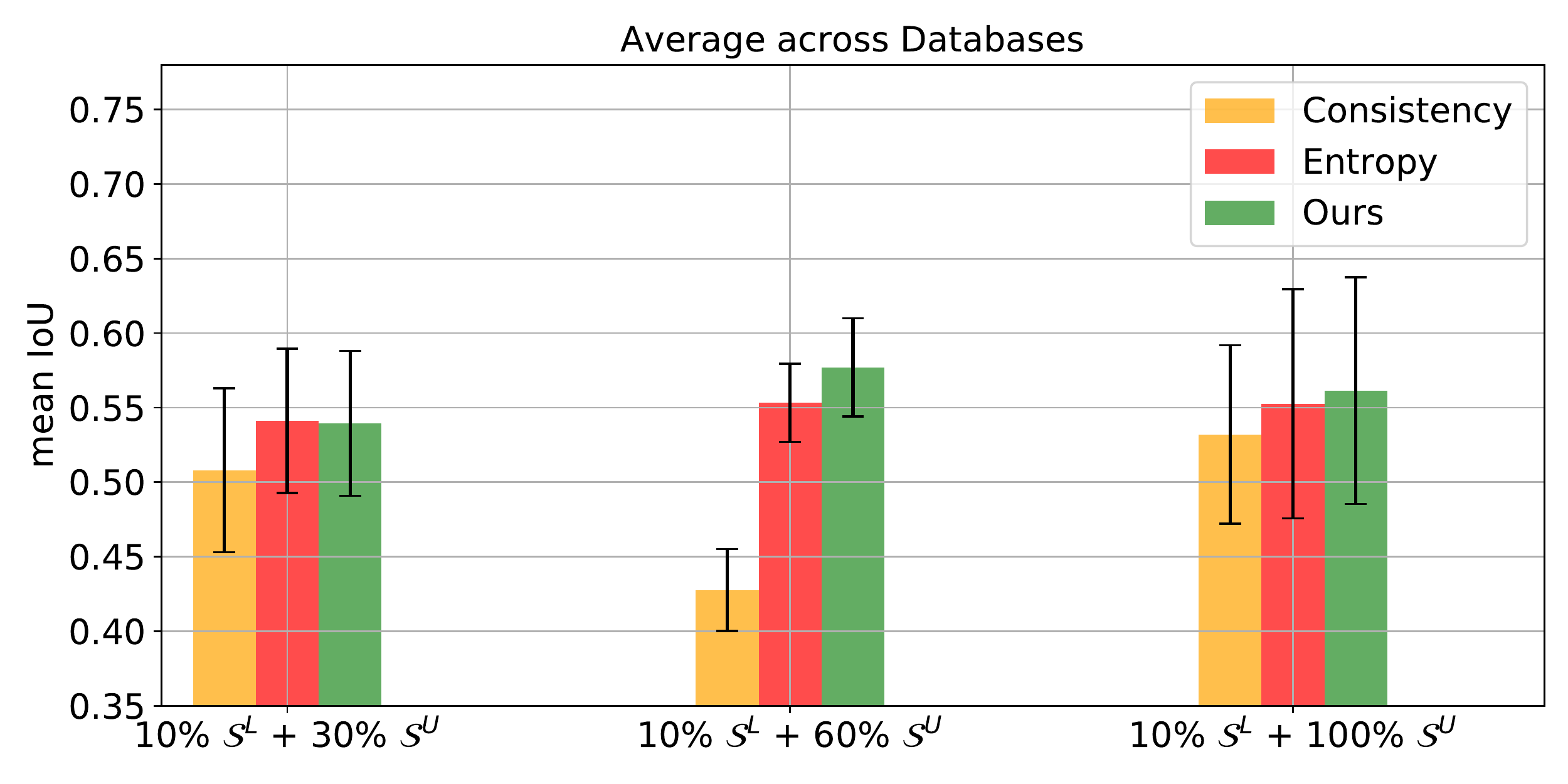}} 
    \subfigure[3-shot.]{\includegraphics[width=0.45\textwidth]{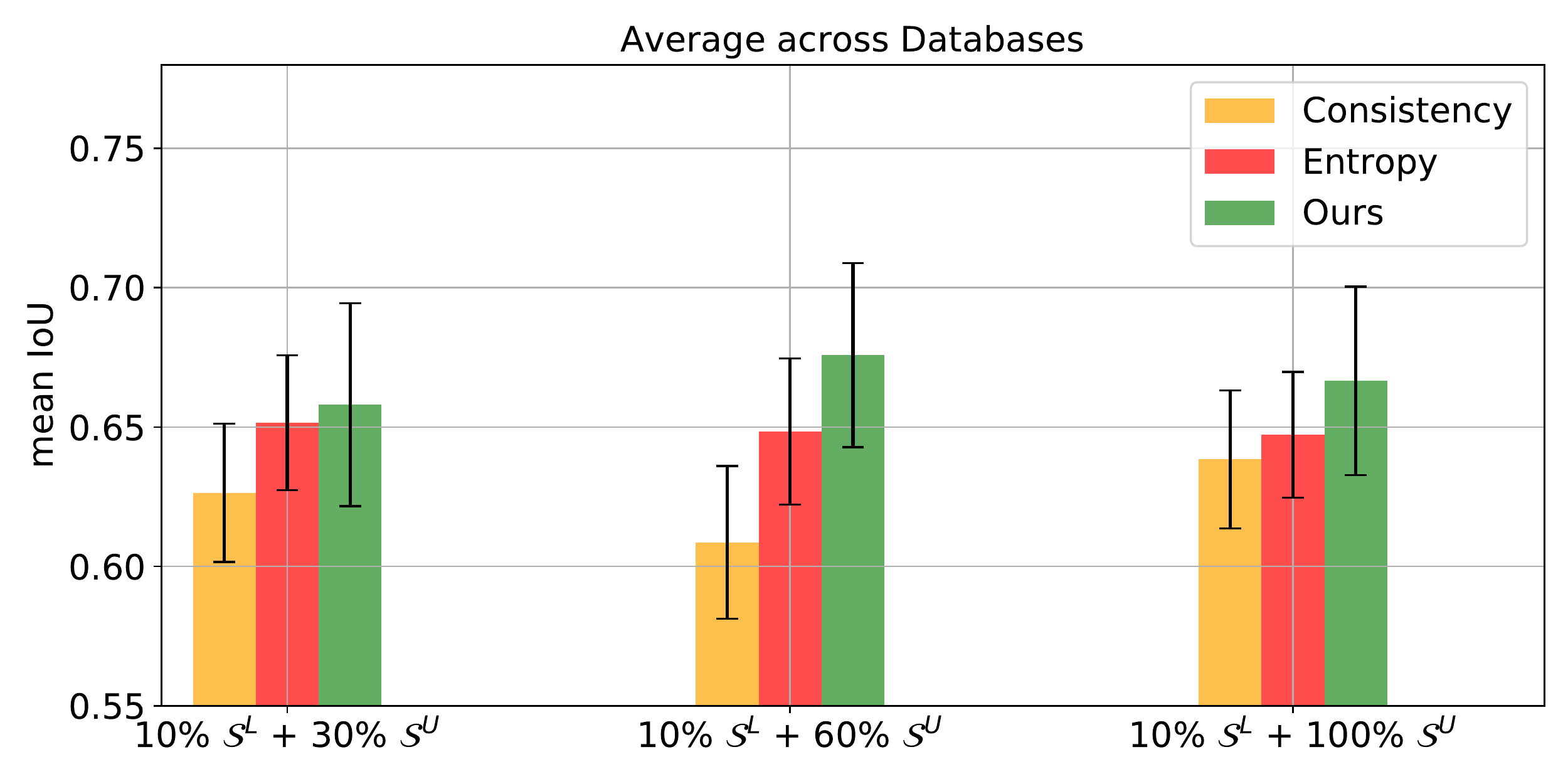}}
   \subfigure[5-shot.]{\includegraphics[width=0.45\textwidth]{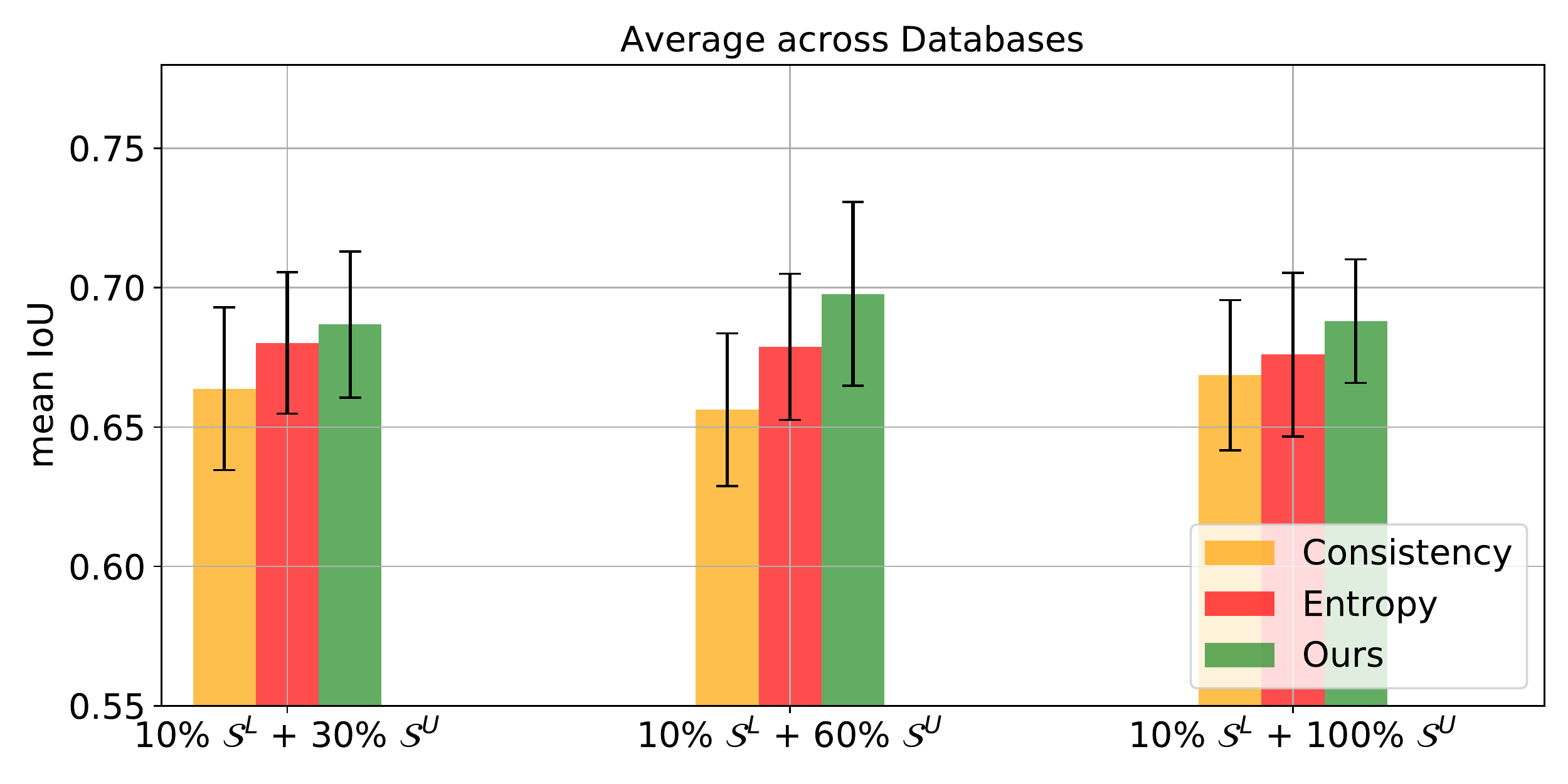}}
   \subfigure[7-shot.]{\includegraphics[width=0.45\textwidth]{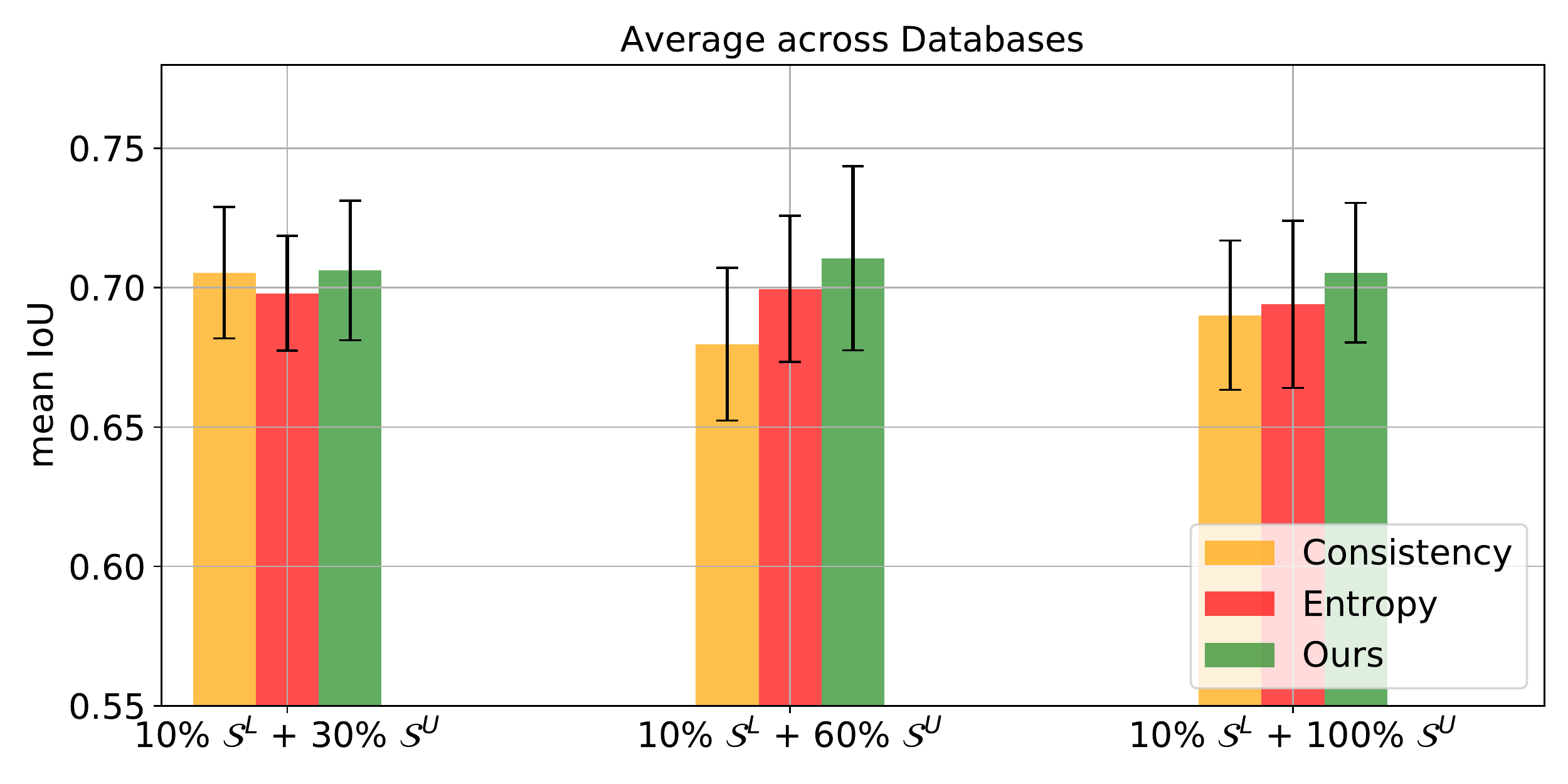}}
   \subfigure[10-shot.]{\includegraphics[width=0.45\textwidth]{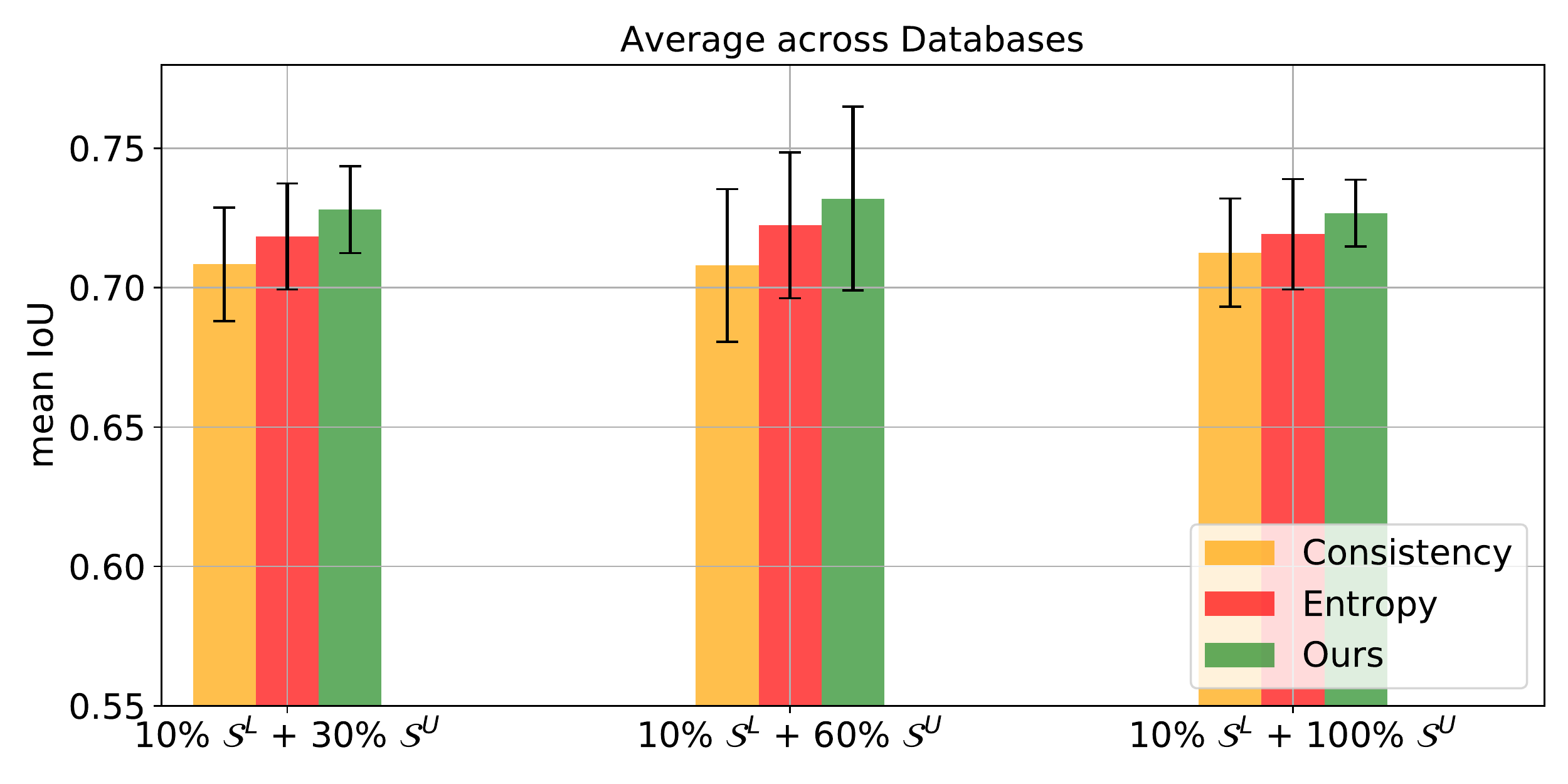}}
   \subfigure[SimCLR, rotation and supervised model against the best result of our approach i.e. $10\% \: \mathcal{S}^{L} + 60\% \: \mathcal{S}^{U}$.]{\includegraphics[width=0.45\textwidth]{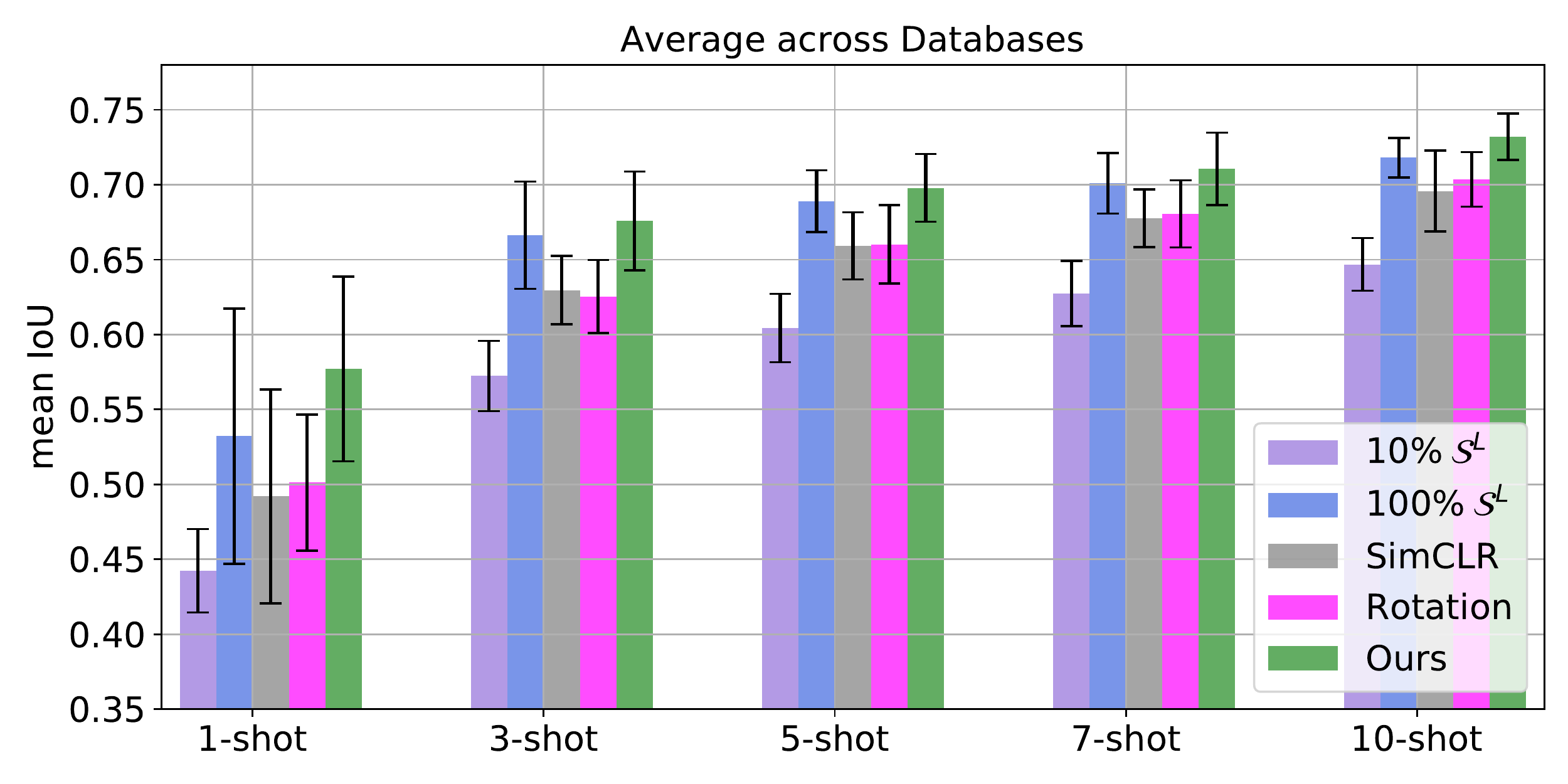}}

    \caption{Mean intersection over union (IoU) comparison using all data sets. We show results of training on $10\% \: \mathcal{S}^{L}$ and $\{30\%$, $60\%$, $100\% \} \: \mathcal{S}^U$ with consistency, entropy and our edge-detection task (Ours). Results are reported for 1 to 10-shot learning. Moreover, we compare the best result of our approach i.e. $10\% \: \mathcal{S}^{L}+60\% \: \mathcal{S}^U$ to SimCLR, rotation and fully-supervised models.}
    \label{fig:ss}
\end{figure}

\begin{center}
\begin{longtable}{|l|l|l|l|l|l|l|} 
\caption{Mean intersection over union (IoU) results for each target dataset using our edge-based learning (\textbf{\textit{Ours}}), consistency \cite{scribble2label} and entropy \cite{grandvalet2005semi} semi-supervised approaches. Also, we show results of SimCLR \cite{chen2020simple}, rotation\cite{gidaris2018unsupervised}, and supervised learning \cite{Dawoud2020}.}
\label{numerical}

\\ \hline
\multicolumn{7}{|c|}{Target: TNBC} \endfirsthead 
\hline
\multicolumn{1}{|c|}{Setting} & \multicolumn{1}{c|}{Method} & \multicolumn{1}{c|}{1-shot} & \multicolumn{1}{c|}{3-shot} & \multicolumn{1}{c|}{5-shot} & \multicolumn{1}{c|}{7-shot} & \multicolumn{1}{c|}{10-shot} \\ 
\hline
100\%~$\mathcal{S}^{L}$ & Supervised & 31.4$\pm$8.2  & 42.4$\pm$2.4 & 44.7$\pm$2.4 & 45.9$\pm$2.4 &  48.5$\pm$1.3\\* 
\hline
\multirow{2}{*}{100\%~$\mathcal{S}^{U}$} & SimCLR & 35.8$\pm$3.7 & 40.2$\pm$2.9 & 42.3$\pm$1.9& 42.6$\pm$2.4  & 45.9$\pm$1.7   \\* 
\cline{2-7}
 & Rotation & 37.0$\pm$3.0 & 41.4$\pm$2.7 & 43.9$\pm$1.8 & 44.4$\pm$2.6 & 48.3$\pm$1.5 \\ 
\hline
10\%~$\mathcal{S}^{L}$ & Supervised & 34.9$\pm$2.9 & 39.3$\pm$2.8 & 41.0$\pm$2.0 &40.9$\pm$2.5  & 44.0$\pm$1.8 \\* 
\hline
\multirow{3}{*}{\quad + 30\%~$\mathcal{S}^{U}$} & Entropy & 37.1$\pm$3.9 & 41.4$\pm$2.4 & 44.8$\pm$1.9 & 45.6$\pm$2.2 & 49.0$\pm$1.6  \\* 
\cline{2-7}
 & Consistency & 37.2$\pm$6.5 & 42.0$\pm$2.1 & 45.5$\pm$1.8 & 46.4$\pm$2.5 & 49.1$\pm$1.7 \\* 
\cline{2-7}
 & \textbf{\textit{Ours}} & 38.6$\pm$4.8 & 43.5$\pm$2.6 & 46.7$\pm$2.3 & 47.1$\pm$2.5  & 49.5$\pm$1.2 \\* 
\hline
\multirow{3}{*}{\quad + 60\%~$\mathcal{S}^{U}$} & Entropy & 37.2$\pm$6.5 & 42.0$\pm$2.1 & 45.5$\pm$1.8 & 46.4$\pm$2.5 & 49.1$\pm$1.7 \\* 
\cline{2-7}
 & Consistency & 28.3$\pm$5.0 & 39.6$\pm$2.6 & 42.8$\pm$2.1 & 43.9$\pm$3.0 & 47.6$\pm$2.0 \\* 
\cline{2-7}
 & \textbf{\textit{Ours}} & 37.8$\pm$6.8 & 43.3$\pm$2.4 & 46.9$\pm$1.7 & 47.1$\pm$2.4 & 50.3$\pm$1.3 \\* 
\hline
\multirow{3}{*}{\quad + 100\%~$\mathcal{S}^{U}$} & Entropy & 39.2$\pm$5.4&
42.7$\pm$1.7 &45.0$\pm$1.7 & 45.9$\pm$2.4 &48.7$\pm$1.3  \\* 
\cline{2-7}
 & Consistency&  36.9$\pm$6.3 &41.9$\pm$2.2 &44.5$\pm$1.6 &45.7$\pm$2.8 &48.2$\pm$1.7\\* 
\cline{2-7}
 & \textbf{\textit{Ours}}  &37.9$\pm$8.5 &43.1$\pm$2.0 &46.1$\pm$1.7 &46.4$\pm$2.1 &49.1$\pm$1.2  \\ 
\hline

\multicolumn{7}{|c|}{Target: EM} \\ 
\hline
\multicolumn{1}{|c|}{Setting} & \multicolumn{1}{c|}{Method} & \multicolumn{1}{c|}{1-shot} & \multicolumn{1}{c|}{3-shot} & \multicolumn{1}{c|}{5-shot} & \multicolumn{1}{c|}{7-shot} & \multicolumn{1}{c|}{10-shot} \\ 
\hline
100\%~$\mathcal{S}^{L}$ &  Supervised & 48.6$\pm$3.0&55.6$\pm$2.3 &58.7$\pm$1.6 &60.9$\pm$1.6 &63.7$\pm$2.3   \\* 
\hline
\multirow{2}{*}{100\%~$\mathcal{S}^{U}$} & SimCLR & 44.0$\pm$3.3 &57.8$\pm$3.3 &
64.1$\pm$3.5 &46.3$\pm$1.8 &51.0$\pm$1.3 \\* 
\cline{2-7}
  & Rotation  & 40.5$\pm$2.1 &54.8$\pm$2.4 &60.6$\pm$2.8 &62.6$\pm$1.7 &
64.7$\pm$3.1   \\ 
\hline
10\%$ \mathcal{S}^{L}$ &Supervised  & 31.5$\pm$1.3 &44.9$\pm$2.1 &
50.2$\pm$2.0 &65.0$\pm$2.6 &66.0$\pm$2.7 \\* 
\hline
\multirow{3}{*}{\quad + 30\%$ \mathcal{S}^{U}$} & Entropy & 42.3$\pm$2.4 &
57.1$\pm$3.0 &60.9$\pm$3.0 &63.3$\pm$2.0 &65.6$\pm$3.1 \\* 
\cline{2-7}
 & Consistency & 33.8$\pm$1.1 &52.6$\pm$2.6&58.9$\pm$2.7 &62.3$\pm$1.7&
 64.4$\pm$2.8  \\* 
\cline{2-7}
 & \textbf{\textit{Ours}}  & 47.8$\pm$2.3 & 59.9$\pm$2.0 &63.0$\pm$1.8 &65.0$\pm$2.6 &66.7$\pm$2.4 \\* 
\hline
\multirow{3}{*}{\quad + 60\%$ \mathcal{S}^{U}$} & Entropy  & 45.1$\pm$2.6 &58.3$\pm$3.3 &61.3$\pm$2.7 &64.5$\pm$3.3& 66.2$\pm$3.3  \\* 
\cline{2-7}
 & Consistency &38.5$\pm$1.6 & 54.9$\pm$3.1 &61.3$\pm$2.4 &63.3$\pm$2.9 &66.3$\pm$2.8  \\* 
\cline{2-7}
 & \textbf{\textit{Ours}}  & 44.5$\pm$2.5 &59.3$\pm$2.3 &63.0$\pm$1.6 &
64.6$\pm$2.8 &66.0$\pm$2.7  \\* 
\hline
\multirow{3}{*}{\quad + 100\%$ \mathcal{S}^{U}$} &   Entropy  & 43.8$\pm$2.2&
57.2$\pm$2.6 &61.3$\pm$2.8 &64.8$\pm$2.5 &66.3$\pm$2.7  \\* 
\cline{2-7}
 & Consistency & 44.5$\pm$1.7 &
58.0$\pm$3.6 &61.9$\pm$2.7 &65.2$\pm$2.6 &66.3$\pm$3.3 \\* 
\cline{2-7}
 & \textbf{\textit{Ours}}  &  43.5$\pm$2.5 &59.2$\pm$2.0&62.7$\pm$2.3 &66.0$\pm$3.4&68.0$\pm$1.8   \\* 
\hline
\multicolumn{7}{|c|}{Target: ssTEM} \\ 
\hline
\multicolumn{1}{|c|}{Setting} & \multicolumn{1}{c|}{Method} & \multicolumn{1}{c|}{1-shot} & \multicolumn{1}{c|}{3-shot} & \multicolumn{1}{c|}{5-shot} & \multicolumn{1}{c|}{7-shot} & \multicolumn{1}{c|}{10-shot} \\ 
\hline
 100\% $\mathcal{S}^{L}$& Supervised & 44.3$\pm$3.2&58.7$\pm$9.9 &60.8$\pm$2.0 &
62.1$\pm$2.4 &63.7$\pm$2.3  \\* 
\hline
\multirow{2}{*}{100\%~$\mathcal{S}^{U}$}& SimCLR &  30.3$\pm$1.3&
49.5$\pm$7.1 &51.7$\pm$1.9 &55.1$\pm$1.8 &55.2$\pm$2.4  \\* 
\cline{2-7}
  &Rotation& 32.3$\pm$1.8&52.3$\pm$2.5 &57.4$\pm$2.2 &59.9$\pm$1.8 &
60.5$\pm$1.8   \\ 
\hline
 10\%$ \mathcal{S}^{L}$&Supervised&25.7$\pm$1.1 &45.5$\pm$6.5 &48.5$\pm$1.5&52.1$\pm$1.4 &
52.5$\pm$1.9 \\* 
\hline
\multirow{3}{*}{\quad + 30\%$\mathcal{S}^{U}$} & Entropy &42.6$\pm$3.5 &
57.3$\pm$2.8 &61.5$\pm$2.4 &63.6$\pm$2.3 &64.0$\pm$2.7 \\* 
\cline{2-7}
 & Consistency &32.0$\pm$2.1 &52.7$\pm$2.7 &57.8$\pm$2.6 &59.7$\pm$2.4 &
60.5$\pm$2.1 \\* 
\cline{2-7}
 & \textbf{\textit{Ours}}  & 46.3$\pm$2.8 &61.7$\pm$9.5 &63.5$\pm$2.3 &
65.2$\pm$2.5&66.6$\pm$2.3    \\* 
\hline
\multirow{3}{*}{\quad + 60\%$\mathcal{S}^{U}$} & Entropy  &  43.3$\pm$3.4 &
58.0$\pm$3.3 &63.0$\pm$2.9&64.6$\pm$2.2&65.7$\pm$3.1  \\* 
\cline{2-7}
 & Consistency &34.2$\pm$2.6 &54.5$\pm$2.8 &59.9$\pm$2.6 &61.4$\pm$2.5 &
62.2$\pm$3.1  \\* 
\cline{2-7}
 & \textbf{\textit{Ours}}  &46.5$\pm$3.0&61.3$\pm$8.6 &63.2$\pm$1.9 &
 64.9$\pm$2.0 &66.5$\pm$1.5   \\* 
\hline
\multirow{3}{*}{\quad + 100\%$\mathcal{S}^{U}$} &   Entropy  &  46.5$\pm$2.8&
57.6$\pm$2.6 &63.0$\pm$2.6&64.1$\pm$2.9 &66.0$\pm$2.6 \\* 
\cline{2-7}
 & Consistency & 43.8$\pm$3.4&58.2$\pm$2.9 &
63.1$\pm$2.4 &64.5$\pm$2.9 &66.0$\pm$2.5 \\* 
\cline{2-7}
 & \textbf{\textit{Ours}}  &  42.1$\pm$2.7 &58.6$\pm$9.4 &61.0$\pm$2.0 &
63.1$\pm$2.4 &64.7$\pm$2.0   \\* 

\hline
\end{longtable}
\end{center}

In Fig.~\ref{fig:ss}a, b, c, d and e, we show the mean IoU of our approach for different percentage of unlabelled and labelled data averaged over all data sets for 1 to 10-shots.  We compare our results with entropy \cite{grandvalet2005semi} and consistency regularization \cite{scribble2label}.
Additionally, in Fig.~\ref{fig:ss}e we compare the best result of our edge-detection task with the results from \cite{Dawoud2020}, which is a supervised model trained using $100\% \: \mathcal{S}^L$. Also, we compare to a supervised model train using $10\% \: \mathcal{S}^L$ as well as SimCLR \cite{chen2020simple} and rotation \cite{gidaris2018unsupervised}. Moreover, we present mean IoU results in Table~\ref{numerical} for each target dataset individually. Our visual results are illustrated in Fig.~\ref{fig:VisualComp1}. Additional visual and numerical results could be viewed in supplementary material.

First, it is clear from Fig.~\ref{fig:ss}a, b, c, d and e that training with our approach outperforms entropy and consistency regularization when fine-tuned to the target using few annotated samples. Second, we notice that the overall performance of our approach is further boosted by using 60\% of $\mathcal{S}^{U}$ across all $K$-shot learning experiments. Nevertheless, the use of 100\% $\mathcal{S}^{U}$ slightly drops the segmentation performance. We attribute this drop as an over-regularization case, where adding more unlabelled data for learning the proxy task negatively impacts the model's performance on the cell segmentation task. Next, it is clear from Fig.~\ref{fig:ss}f that training a fully-supervised model only on $10\% \: \mathcal{S}^{L}$ results in poor performance relative to all other approaches. Furthermore, we notice that by utilizing 60\% of $\mathcal{S}^{U}$ for learning our proposed edge-detection task jointly with $10\% \: \mathcal{S}^{L}$, we achieve far better results than a fully-supervised model. Finally, we observe that relying on SimCLR and rotation as self-supervised tasks lags behind our proposed approach. In Table~\ref{numerical}, we observe mostly better performance across each target dataset using our approach compared to the proposed semi-supervised, self-supervised and supervised baselines. Although, we notice at few cases slightly better results using consistency and entropy regularization. We argue that our proposed proxy task is more related to the microscopy image cell segmentation problem, hence, we demonstrate an overall better performance suitable for microscopy image datasets of different image types and cell segmentation task.

\section{Conclusion}

We proposed to form edge-based maps for unlabelled microscopy images to self-supervise the training of a deep neural network. In addition, a small number of training samples containing ground-truth segmentation was used for learning to segment cells. In our evaluations, we show that training with our proxy task and only 10\% of the annotated training sets achieves equal performance to a fully supervised approach, which dramatically reduces time and cost of experts to annotate cells in microscopy images. Moreover, we reach a better performance than the related works on semi-supervised learning and self-supervised learning. 

\section*{Acknowledgments}
This work was partially funded by Deutsche Forschungsgemeinschaft (DFG), Research Training Group GRK 2203: Micro- and nano-scale sensor technologies for the lung (PULMOSENS), and the Australian Research Council through grant FT190100525.  G.C. acknowledges the support by the Alexander von Humboldt-Stiftung for the renewed research stay sponsorship.

%
% ---- Bibliography ----
%
% BibTeX users should specify bibliography style 'splncs04'.
% References will then be sorted and formatted in the correct style.
%
\bibliographystyle{splncs04}
\bibliography{paper0005.bib}

\end{document}